\def\year{2021}\relax
\documentclass[letterpaper]{article} 
\usepackage{aaai21}  
\usepackage{times}  
\usepackage{helvet} 
\usepackage{courier}  
\usepackage[hyphens]{url}  
\usepackage{graphicx} 
\usepackage{natbib}  
\usepackage{caption} 
 \usepackage{balance} 
\usepackage{url}
\usepackage{float}
\usepackage{amsmath,graphicx,amsfonts,amssymb,url}
\usepackage{setspace}
\usepackage[caption=false]{subfig}
\usepackage{blindtext}
\usepackage[inline]{enumitem}
\usepackage{color}
 \usepackage{subfig}
\usepackage{multirow}
\newsavebox{\measurebox}
\usepackage{minibox}
\usepackage{graphicx}
\definecolor{mypink1}{rgb}{0.858, 0.188, 0.478}
\usepackage{csquotes}

\usepackage{algorithm}
\usepackage{algorithm,algorithmic,amsmath}

\newcommand{\comment}[1]{}

\newcommand{\squeezeup}{\vspace{-2.4mm}}
 \usepackage{ulem}
 
\title{Strategy for Boosting Pair Comparison and Improving Quality Assessment Accuracy}
\author{
Suiyi Ling, \textsuperscript{\rm 1}
Jing Li, \textsuperscript{\rm 2}
Anne Flore Perrin, \textsuperscript{\rm 1}
Zhi Li, \textsuperscript{\rm 3}
Lukáš Krasula, \textsuperscript{\rm 3}
Patrick Le Callet\textsuperscript{\rm 4} 
    \\
}
\affiliations{
 \textsuperscript{\rm 1} CAPACITÉS 
 \textsuperscript{\rm 2} Alibaba Group 
 \textsuperscript{\rm 3} Netflix  
 \textsuperscript{\rm 4} University of Nantes
    \\
   suiyi.ling@capacites.fr,  jing.li.univ@gmail.com, Anne-Flore.Perrin@capacites.fr, \\
   zli@netflix.com, 
   lkrasula@netflix.com, 
   patrick.lecallet@univ-nantes.fr
}
    
\begin{document}

\maketitle

\begin{abstract}

 The development of rigorous quality assessment model relies on the collection of reliable subjective data, where the perceived quality of visual multimedia is rated by the human observers. Different subjective assessment protocols can be used according to the objectives, which determine the discriminability and accuracy of the subjective data. 
 Single stimulus methodology, \textsl{e.g.,} the Absolute Category Rating (ACR) has been widely adopted due to its simplicity and efficiency. However, Pair Comparison (PC) is of significant advantages over ACR in terms of discriminability. In addition, PC avoids the influence of observers' bias regarding to their understanding of the quality scale. Nevertheless, full pair comparison is much more time consuming. In this study, we therefore 1) employ a generic model to bridge the pair comparison data and ACR data, where the variance term could be recovered and the obtained information is more complete; 2) propose a fusion strategy to boost pair comparisons by utilizing the ACR results as initialization information; 3) develop a novel active batch sampling strategy based on Minimum Spanning Tree (MST) for PC. In such a way, the proposed methodology could achieve the same accuracy of pair comparison but with the compelxity as low as ACR. Extensive experimental results demonstrate the efficiency and accuracy of the proposed approach, which outperforms the state of the art approaches. 
 

 
\end{abstract}

\section{Introduction}
\label{sec:intro}

Recently, with the prosperity of multimedia technologies and the popularization of high-quality contents, users are becoming increasingly quality-aware~\cite{moldovan2013user}. To catch up with the growing expectation of higher quality-of-experience, robust quality metric that is of higher discriminability, especially for higher-quality contents with less visual difference, is in urgent need~\cite{nandakumar2019accuracy}. Due to the `range effect', higher capability of distinguishing pairs in narrow quality range~\cite{krasula2017quality_highligh} is essential. The improvement of objective quality models depends on the accuracy and the discriminability of the subjective data collected utilizing a certain subjective quality protocol or methodology from human observers. According to the standards and recommendations~\cite{itu1999methods,sector2012recommendation} published for multimedia quality assessment, the subjective quality protocols could be classified into two main categories including the rating and the comparative methodologies. Absolute Category Rating (ACR) is one of the most commonly utilized single stimulus rating protocols, while Pair Comparison (PC) is the most widely employed comparative approach~\cite{perez2019pairwise}. 

On one hand, single stimulus rating methods are suitable when the stimuli are easy to be distinguished. Regardless of their simplicity and efficiency, they are prone to objects' bias and inconsistency~\cite{li2020simple} as observers may have different interpretation of the quality scale~\cite{li2020gpm}, memory ability ~\cite{le2016influence}, and the task's difficulty varies~\cite{lakshminarayanan2013inferring}, \textsl{etc.} On the other hand, pair comparison has its own advantage of discriminability, as the preference of the observer between each pair of stimuli is asked instead of a score of an individual stimulus in a discrete or continuous scale. Although full pair comparison is of advantages in distinguishing stimuli with small visual difference, it is time consuming, since the number of comparisons increases exponentially with the increase of the number of stimuli. Thus, better strategy is required to achieve a better trade-off between the discriminability and the efficiency of the subjective protocol.

 The emergence of crowdsourcing has sparked a lot of interest for the quality assessment community. There is a plethora of pairwise comparison experiments that were conducted via crowdsourcing platforms~\cite{xu2017hodgerank}. Aiming at obtaining crowdsourcing ranking more efficiently and accurately by making better use of the information of past labeled pairs, many~\textsl{active sampling} strategies were purposed to boost the accuracy of ranking aggregation~\cite{pfeiffer2012adaptive,li2018hybrid}. Since most of the existing subjective studies were conducted using single stimuli methods, the state-of-the-art active sampling strategies are of great potential to be exploited to boost the accuracy and discriminability of subjective data collected with rating protocols.

 Recall that after the collection of pair comparison subjective data, models like the Bradley-Terry (BT) or the Thurstone-Mosteller (TM), \textsl{i.e.}, the Thurstone Case V , are commonly adapted to convert the pair comparison results into quality scores or ranking. Thus, they are also the fundamental cornerstones of the subjective data collection procedure. However, most of the existing standardized conversion models neglect the variance of subjective data, and thus may lose important information including the task difficulties, observers' biases and inconsistencies, \textsl{etc.}
 
 
 In this study, a novel framework is presented to boost the  pair comparison with ACR data so that the quality assessment accuracy could be further improved. The contributions of this framework are threefold: 
 
 \begin{itemize}
     \item A brand-new fusion scheme that combines the ACR and PC data to achieve better trade-off between the accuracy and efficiency of the subjective data collection procedure.
     \item We adapt the Thurstone Model Case III for pairwise comparison data conversion, where the variance of stimuli could be recovered. By doing so, we narrow the gap between the ACR and PC data and avoid relevant information loss regarding the variance.
    \item  A new version of Hybrid-MST, where the active batch sampling strategy is strengthen by the ACR initialization and novel recovering model. 
 \end{itemize}


\section{Related work}
 \textbf{Pairwise preference aggregation/conversion model:} In the past decades, many models have been proposed to covert or aggregate the pair comparisons responses to rating or ranking scale. The heuristic approach of Emerson~\textsl{et al.}~\cite{emerson2013original}, and the probabilistic permutation based models~\cite{plackett1975analysis} are typical examples. In addition, the Thurstone-Mosteller~\cite{mosteller2006remarks} and Bradley Terry~\cite{bradley1952rank} are another two widely used linear models of paired comparisons, where the probabilities of preference between stimuli are converted to scales. Due to the issues of computation-complexity or parameter-estimation, several models were developed to improve the the parameter-optimization procedure~\cite{azari2012random,lu2011learning}. For instance, a generalized method-of-moments was presented~\cite{soufiani2013generalized} to speed up existing model with well-designed generalized moment conditions. In~\cite{freund2003efficient}, the RankBoost was proposed for combining multiple preferences. Shah \textsl{et al.} introduced the min-max bounds on the optimal error~\cite{shah2016estimation} to improve the parametric ordinal models. Other type of conversion models were proposed based on inferring the underlying latent scores~\cite{dangauthier2008trueskill,wauthier2013efficient}. Among the existing models, there are only few of them consider fusing the rating score with the comparison subjective data. The relationship between the rating and pairwise comparison data was studied in~\cite{watson2001measurement}. A unified probabilistic model was presented in~\cite{ye2014active} to aggregate rating scores and pairwise comparisons subjective results. Yet none of these models seek to recover the variance of the stimuli. In one of the most recent study~\cite{perez2019pairwise} a Thurstone Case V based probabilistic model was proposed to combine the rating and comparison subjective data, but no active sampling strategy was considered.

\textbf{Sampling strategy for pair comparison:} To infer the ranking from pair comparison data, a significant number of pairs are required to be compared. Since data sampling is one of the simplest way to reduce the cost of pairwise labeling, random sampling strategies, \textsl{e.g.,} the model proposed by Dykstra~\textsl{et al.}~\cite{dykstra1960rank}, were developed in earlier studies. The HodgeRank on Random Graph (HRRG)~\cite{lin2012hodgerank} was developed based on random graph theory and Hodge decomposition of the graphs  paired comparison preferences. An Adaptive Rectangular Design (ARD) was shown in~\cite{li2013boosting}, to sample novel pairs based on the predicted ranks using current pair comparison results. As active learning has been established as an effective approach for many domains, it is also adopted to improve the performance of pair comparison aggregation. Jamieson \textsl{et al.} proposed an active ranking recovery model by embedding objects into a $d$-dimensional Euclidean space~\cite{jamieson2011active}. In~\cite{pfeiffer2012adaptive}, a Bayesian optimization scheme was proposed based on TM model. Similarly, the Crowd-BT~\cite{chen2013pairwise} model was proposed following a similar concept but using BT model instead. The HRRG was improved in~\cite{xu2017hodgerank} by maximizing information gains of pairs. Recently, a Hybrid active sampling strategy was proposed by Li~\textsl{et al.}~\cite{li2018hybrid}, where a batch mode was designed using the Minimum Spanning Tree (Hybrid-MST) for the ranking of information gains. It was proven in~\cite{li2018hybrid} that Hybrid-MST achieves best aggregation performance compared to the other state-of-the-art models, and is of significant advantages in terms of efficiency when utilizing the batch mode for parallel labeling on the crowd sourcing platforms. 
However, none of them consider to boost the existing subjective data collected via single stimulus protocols.



\section{The Proposed Framework\footnote{The source code and the table that summarizes all the variables used in this paper are provided in the supplemental material.}}
Even though the ACR test may fail to accurately rank two stimuli with enough precision or discriminability compared to pair comparisons due to a series of factors introduced in previous sections, it can provide a coarse estimation of the underlying quality. Since the underlying ground truth of ACR and the PC test is consistent, complete pair comparisons become unnecessary once the coarse estimation is available. Therefore, resources could be spent on more informative pairs to obtain finer discrimination on pairs with similar quality and high uncertainty. Our framework is inspired by this idea, details are described below.

\subsection{0. Problem setup and overview of the framework}
Let us assume that we have $n$ test stimuli $A_{1},A_{2}, ...A_{n}$ in a pairwise comparison experiment. The underlying quality scores of these objects are $\mathbf{s} = (s_{1}, s_{2},...s_{n})$. In addition, each test stimulus has its task difficulty, $\sigma_{i}$, which determines the participant's perceptual opinion diversity, \textsl{i.e.}, higher $\sigma_{i}$ indicates people's opinion are more diverse, lower $\sigma_{i}$ represents opinion consistency. Thus, the quality character of a test stimulus $A_{i}$ can be described by a Gaussian distribution $\mathcal{N}(s_i, \sigma_{i}^{2})$.

\begin{figure}[!htbp]
    \centering
    \includegraphics[width=\columnwidth]{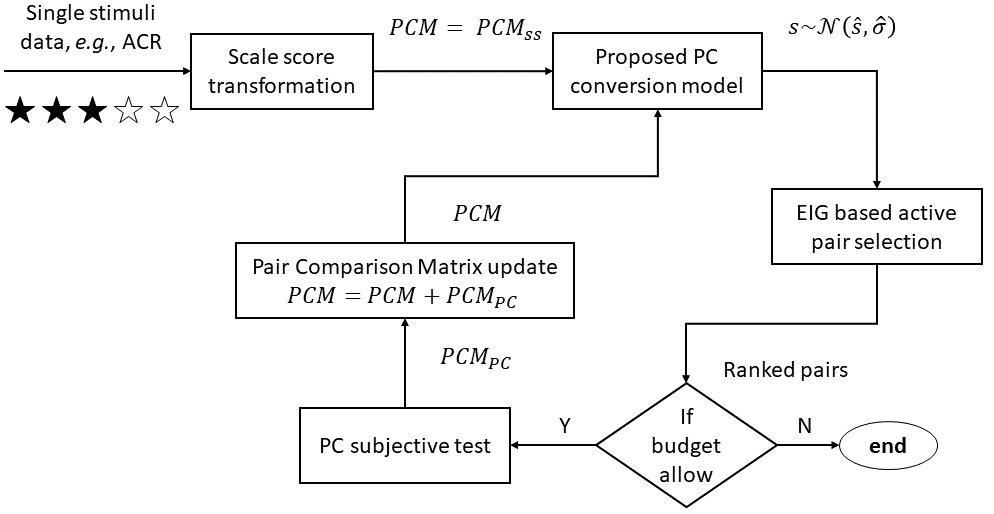}
    \caption{Diagram of the overall framework. \squeezeup  }
    \squeezeup    
    \label{fig:OF} 
\end{figure}

The diagram of the proposed framework is summarized in Figure~\ref{fig:OF}. In a nutshell, 1) given the subjective data collected from any single stimuli test/tests (\textsl{e.g., ACR}), the linear scores are first transformed into an initial pair comparison matrix $PCM_{SS}$, and the overall pair comparison matrix is initialized by $PCM = PCM_{SS}$; 2) Afterwards, the proposed pair comparison data conversion model is applied to approximate prior information on $\mathcal{N}(\hat{s},\hat{\mathbf{\sigma}}^{2})$, where $\mathbf{\hat{s}} = (\hat{s_{1}}, \hat{s_{2}},...\hat{s_{n}})$ is the approximated/recovered  underlying quality scores regarding $\mathbf{s}$, and $\hat{\mathbf{\sigma}} = (\hat{\sigma_1}, \hat{\sigma_2},...,\hat{\sigma_n})$ is the recovered underlying standard deviation \textsl{w.r.t.} $\mathbf{\sigma}$; 3) With the recovered $\hat{s}, \hat{\sigma}$, state-of-the-art active sampling strategy, \textsl{e.g.,} the Hybrid-MST~\cite{li2018hybrid}, is adapted to select the most informative pairs. Then, the pairs with the highest information gains are selected for pair comparison subjective test to collect an extract set of PC data $PCM_{PC}$ with $n_{pc}$ pairs. $n_{pc}$ is the number of pairs decided by the total budget of the subjective test. By doing so, the discriminability and reliability of the PC data could be enhanced by considering only the most valuable pairs. Moreover, the complexity/cost of the experiment is reduced when compared to full pair comparison as only the top $n_{pc}$ pairs are compared instead of $n(n-1) /\ 2$; 4) Then  the overall $PCM$ is updated by adding the new $PCM_{PC}$. Procedure (1-4) is repeated until reaching the total budget of the subjective test. Let $n_{itr}$ be the number of iteration from step 1 to 4 mentioned above, then the total number of pairs compared in the test equals to $ n_{budget} = n_{pc} \times n_{itr}$. 



\subsection{1. PCM initialization: Scale score transformation}    
The process of transforming the linear scale scores collected from a single stimuli subjective experiments to pair comparison matrix is described in \textbf{Algorithm 1}. $n_{obs}$ denotes the number of total observers, $r_i^{obs}$ is the individual rating of stimulus $A_i$ from the $obs^{th}$ observer. During the procedure, if the observer rates $A_i$ over $A_j$, then the corresponding $(i,j)^{th}$ element within the initialized pair comparison matrix $PCM_{SS}$ accumulates 1, and vice versa. For the remaining pairs, where no preference is given, 0.5 is assigned.  The output of this procedure is considered as the initialized pair comparison matrix, \textsl{i.e.,} $PCM_{(itr=1)} = PCM_{SS}$, of the proposed boosting framework, where $itr=1$ indicates the first iteration. 
 
\begin{algorithm}[!hpbt]
\begin{algorithmic} 
\caption{Procedure of transforming linear subjective score into pair comparison matrix. }
\STATE  \textbf{for} $obs \in \{ 1, \cdots, obs , \cdots,   n_{obs} \}$ \textbf{do}  
  \STATE \ \ \ \  \textbf{for} $ A_i  \in   \{ 1, \cdots, A_i , \cdots, A_n \}$  \textbf{do}   
\STATE \ \ \ \ \ \ \ \     \textbf{for} $A_j  \in   \{ 1, \cdots, A_j , \cdots, A_n \}$  \textbf{do}   
\STATE      \ \ \ \ \ \ \ \ \ \ \ \ \ \   \textbf{if}  $r_i^{obs} > r_j^{obs} $ \textbf{then}   
\STATE  \ \ \ \ \ \ \ \ \ \ \ \  \ \ \ \ $ PCM_{SS}(i,j) = PCM_{SS}(i,j) + 1 $
\STATE       \ \ \ \ \ \ \ \ \ \ \ \ \ \   \textbf{elseif} $r_i^{obs} < r_j^{obs} $ \textbf{then}            
\STATE  \ \ \ \ \ \ \ \ \ \ \ \  \ \ \ \ $ PCM_{SS}(j,i) = PCM_{SS}(j,i) + 1 $
\STATE      \ \ \ \ \ \ \ \ \ \ \ \ \ \   \textbf{else}           
\STATE  \ \ \ \ \ \ \ \ \ \ \ \  \ \ \ \  $ PCM_{SS}(i,j) = PCM_{SS}(i,j) + 0.5 $
\STATE  \ \ \ \ \ \ \ \ \ \ \ \  \ \ \ \ $ PCM_{SS}(j,i) = PCM_{SS}(j,i) + 0.5 $
 \label{alg:acm_pcm}
 \end{algorithmic} 
\end{algorithm}
  \squeezeup

\subsection{2. Bridge ACR and PC: the proposed PC conversion model}
 
In a pairwise comparison experiment, the annotator's observed score for stimulus $A_{i}$ is $r_{i}$, for stimulus $A_{j}$ is $r_{j}$. If $r_{i} > r_{j} $, then we consider that the annotator prefers stimulus $A_{i}$ over $A_{j}$. Otherwise, the preference is opposite. When we observe $r_{i} = r_{j}$, there is no significant difference between the two candidates. Then, we consider that the annotator makes a random selection.

The observed value $r_{i} - r_{j}$ is determined not only by the two Gaussian distribution $\mathcal{N}(s_i, \sigma_{i}^{2})$ and $\mathcal{N}(s_j, \sigma_{j}^{2})$, but also by the comparison interaction terms. That is to say, in a typical ACR test, the two observed quality scores for $A_i$ and $A_j$ are independent. However, during the comparison procedure, they are not independent any more. The whole pair comparison procedure can be modeled as follows: 

 
\begin{equation}
r_{i} - r_{j} \sim \mathcal{N}(s_i-s_j, \sigma_i^2+\sigma_j^2-2\sigma_{ij}),
\end{equation}
where $\sigma_{ij}$ is the co-variance term. The probability of selecting $A_i$ over $A_j$ is denoted as $Pr(A_{i} \succ A_{j})$, which can be calculated by:
\begin{equation}
Pr(A_{i} \succ A_{j}) = \Phi \left(\frac{s_i-s_j}{\sqrt{\sigma_i^2+\sigma_j^2 - 2 \sigma_{ij} }}\right), 
\end{equation}

where $\Phi(x)=\frac{1}{\sqrt{2\pi}}\int_{-\infty}^{x}e^{-\frac{t^{2}}{2}}dt$
is the cumulative function of a Gaussian distribution with $N(0,1)$.

\subsubsection{A Generalized Pair Comparison Model}
Ideally, we should estimate the aforementioned parameters through the pairwise comparison observations. However, in this case, the number of parameters is much larger than the number of observations, which makes the equation to have an infinite number of solutions. To resolve this, we abandon the interaction term under the assumption that the influence of the interaction term is limited when compared with the sum of $\sigma_i^2$ and $\sigma_j^2$. The model is then defined as follows, which is in fact the Thurstone Model Case III~\cite{thurstone1927law}:
\begin{equation}
Pr(A_{i} \succ A_{j}) = \Phi \left(\frac{s_i-s_j}{\sqrt{\sigma_i^2+\sigma_j^2}}\right)
\end{equation}

\subsubsection{Maximization Likelihood Estimation (MLE) Procedure \footnote{More details of the MLE procedure, \textsl{e.g.,} the calculation of relevant derivatives regarding the utility function, could be found in the supplemental material. }}
To infer the $2n$ parameters of Thurstone model Case III , the Maximum Likelihood Estimation (MLE) method is adopted in this study. Given the pairwise comparison results arranged in a matrix $\mathbf{M} = (m_{ij})_{n\times n}$, where $m_{ij}$ represents the total number of trial outcomes $A_{i} \succ A_{j}$, the likelihood function takes the shape:
\begin{equation}
L(\mathbf{s}|\mathbf{M}) = \prod_{i<j}\pi_{ij}^{m_{ij}}(1-\pi_{ij})^{m_{ji}}
\end{equation}
Replacing $\pi_{ij}$ by $\Phi \left(\frac{s_i-s_j}{\sqrt{\sigma_i^2+\sigma_j^2}}\right) $, and maximizing the log likelihood function $logL(\mathbf{s}|\mathbf{M})$, we could obtain the MLEs $\hat{\mathbf{s}} = (\hat{s_1},\hat{s_2},...,\hat{s_n})$, $\hat{\mathbf{\sigma}} = (\hat{\sigma_1}, \hat{\sigma_2},...,\hat{\sigma_n})$. 

To obtain the confidence intervals of the MLEs, the second-order derivatives should be calculated and the Hessian matrix $H$ can be constructed. For $\mathbf{s}$, we have:

\begin{equation}
H=\begin{bmatrix}
\frac{\partial ^{2}logL}{\partial s_{1}^{2}} &\cdots &\frac{\partial ^{2}logL}{\partial s_1\partial s_n} \\ 
\cdots&\ddots   & \cdots \\ 
\frac{\partial ^{2}logL}{\partial s_n\partial s_1}  & \cdots & \frac{\partial ^{2}logL}{\partial s_{n}^{2}} 
\end{bmatrix}
\end{equation}
Following \cite{wickelmaier2004matlab}, we construct a matrix C, which has the following form by augmenting the negative $H$ a column and a row vector of ones and a zero in the bottom right corner:
\begin{equation}
C=\begin{bmatrix}
-\bf{H} & \bf{1}\\ 
\bf{1}' & 0 
\end{bmatrix}^{-1}
\end{equation}
The first $n$ columns and rows of $C$ form the estimated covariance matrix of $\hat{\mathbf{s}}$, \textsl{i.e.}, $\hat{\Sigma}$. Similar procedure can be implemented for the calculation of covariance matrix of $\hat{\sigma}$.

\subsection{3. Boosting Procedure: Expected Information Gain (EIG) based active pair selection}

In order to recover the underlying rating of the stimuli from the sparse and noisy pair comparison subjective data, an active sampling strategy for pairwise preference aggregation was proposed by Li~\textsl{et al}~\cite{li2018hybrid}. Since this model achieves state-of-the-art performance, it is hence adapted in this study to boost the accuracy of non-full pair comparison subjective test. Similarly, we define the utility function as:
\begin{equation}
\begin{array}{ll}
    U_{ij} & =   E(p_{ij}log(p_{ij})) + E(q_{ij}log(q_{ij}))   \\
     & - E(p_{ij})log(E(p_{ij})) - E(q_{ij})log(E(q_{ij})),  \\
     \label{eq:utility_func}
\end{array}
\end{equation}
Differently, in this study,  we have $p_{ij}=\Phi\left(\frac{ s_{i}- s_{j}}{\sqrt{\sigma_{i}^{2}+\sigma_{j}^{2}}}\right)$ and $q_{ij} = 1 - p_{ij}$ as defined in previous section.

For simplicity, we replace $s_i-s_j$, $\hat{s_i}-\hat{s_j}$ with $s_{ij}$ and $\hat{s_{ij}}$ respectively in the remaining of the paper. 

The first term of equation (\ref{eq:utility_func}) could be developed in:

\begin{equation}
\begin{array}{ll}
 E(p_{ij}log(p_{ij}))   & =\int p_{ij} log(p_{ij}) p(s_{ij} ) d  s_{ij}\\
& = \int \Phi(\frac{s_{ij}}{\sqrt{\sigma_{i}^{2}+\sigma_{j}^{2}}}) log \left( \Phi(\frac{s_{ij}}{\sqrt{\sigma_{i}^{2}+\sigma_{j}^{2}}}) \right) \\
& \cdot \frac{1}{\sqrt{2\pi}\sigma_{ij}}e^{-\frac{( s_{ij} -  \hat{s_{ij}} )^2}{2\sigma_{ij}^2}} d s_{ij} .
\end{array}
\end{equation}

By operating the following change of variable: 
\begin{equation}
   x = \frac{s_{ij}  - \hat{s_{ij}} }{\sqrt{2} \sigma_{ij}}  
     \Leftrightarrow s_{ij}  = \sqrt{2} \sigma_{ij}x + \hat{s_{ij}} , 
\end{equation}

we can then obtain a new expression of the first term (same for other terms) of equation (\ref{eq:utility_func}) as being 
\begin{equation}
  E(p_{ij}log(p_{ij})) = \int \frac{1}{\sqrt{\pi}} e^{-x^2}h(\sqrt{2} \sigma_{ij} x + \hat{s_{ij}}) dx
\end{equation}

 In this forms, the Gaussian-Hermite quadrature could be applied to approximate each term by
\begin{equation}
  \sum_{i=1}^N \frac{1}{\sqrt{\pi}} w_i h(x_i).
\end{equation}


    
    
    
%
 
\subsection{4. Information fusion of SS and PC tests:}
After conducting the pair comparison with the selected most informative pairs, a sparse pair comparison matrix $PCM_{PC}$ could be obtained. Therefore the current $PCM_{(itr=i)}$ of the $i^{th}$ iteration is updated via:
\begin{equation}
    PCM_{(itr=i)} = PCM_{(itr=i-1)} + PCM_{PC}.
\end{equation}

\section{Experiment and Analysis ~\footnote{More experimental results and the calculation of the EIG based on TM model are given in the supplemental material.}}
\subsection{Experimental Setup}

\subsubsection{Performance evaluation} The performances of the considered models are estimated by calculating the Spearman's Rank Correlation Coefficient (SROCC) between the ground truth and  obtained estimated scores. Due to limited space, only SROCC are shown in the paper, other results are reported in the supplemental material. Since BT and TM are the most commonly used models, we mainly compared to them during performance evaluation.

\subsubsection{Experiments on simulated data}
A Monte Carlo simulation is conducted on 60 stimuli whose scores are randomly selected from a uniform distribution on the interval of [1 5] with noise $\epsilon_{n} $, which is uniformly distributed between 0 and 0.7 as done in~\cite{li2018hybrid} to simulate the procedure of rating from observers. During the simulation, if the sampled score (from the uniform distribution with noise) $r_i > r_j$, then we consider that $A_i$ is preferred over $A_j$. For statistically reliable evaluations, the simulation experiment was conducted 100 times and the averaged performance are reported. In each iteration, 50 standard trial numbers are simulated (\textsl{i.e.} 50 simulated annotators to compare all $n(n-1)  /\ 2$ pairs using the active learning scheme for pairs sampling). To compare the performances, SROCC is calculated between the simulated ground truth and the estimated scores.

\subsubsection{Experiments on real-world datasets}
In this study, four datasets equipped with both linear quality scores, \textsl{e.g.} MOS obtained using ACR, and the pair comparison ground truth are considered for the performance evaluation of the proposed model. It has to be emphasized that, for the pair comparisons data from the real-world data, only comparisons among PVS from the same contents are available with few cross-content comparison pair in certain datasets. Details of the datasets are summarized below. As there is no real underlying ground truth for the real-world datasets, the results obtained by all observers are considered as the ground truth and the SROCC between it and the estimated scores is calculated for performance estimation. Similar to the simulation test, the experiments were repeated 100 times to simulate the procedure of rating within the active sampling framework, with 50 standard trial numbers per iteration.

\begin{itemize}
	\item \textbf{The DIBR Image dataset:} To quantify the impacts of the Depth Image-Based Rendering (DIBR) algorithms on the perceived quality of free-viewpoint videos, Bosc \textsl{et al.}~\cite{bosc2011towards} have conducted a subjective studies using the ACR and the PC protocols. Three free-viewpoint contents were synthesized using seven different DIBR methods to obtain four different virtual videos corresponding to each reference, which ends out to 84 synthesized free-viewpoint videos. 43 observer participated in the subjective their study for both ACR and PC test.
	
	\item \textbf{The Kaist dataset:}  This dataset was released for studying the influence of visual discomfort, especially motion, on visual experience~\cite{jung2013predicting,li2018exploring}. It contains 36 of the video sequences labeled with both ACR scores and PC preferences. There are 4 motions types including the vertical planar motion, horizontal planar motion, in-depth motion and  the combinations of the three previous motions. During the ACR test, 17 observers were asked to rate the sequence with visual comfort scores (5-point scale values).  In the PC test, totally 180 pairs were collected with 40 naive observers using the same stimuli. 
    
	\item \textbf{The IVC image dataset:} It is one of the earliest and most famous~\cite{ninassi2006pseudo} image quality assessment. Unlike the other famous LIVE image quality assessment database~\cite{sheikh2005live}, it provides both the MOS and the standard deviation of the raw subjective scores, which makes the development of the variance recovery possible. Within the dataset, 10 original images were used, altogether 235 degraded images were obtained via 4 different distortion processes. The original IVC image dataset contain only linear quality scores. Therefore, we also considered the PC dataset summarized in \cite{xu2017hodgerank}, which is composed of 43,266 paired comparisons using images from both the IVC and LIVE datasets. There was altogether 328 observers in the subjective test. Similarly, as there are no ground truth standard deviation of the raw subjective data from the LIVE dataset, we kept only the pairs from the IVC dataset. 
	
 \item   \textbf{The streaming video dataset:} To evaluate how the proposed model could be used for mainstream streaming platform, we have collected 3 contents, \textsl{i.e.,} the Hypothetical Reference Circuits (HRC), from one of the most popular streaming platforms, and the contents were proceed with 4 encoding resolutions (\textsl{i.e.}, 4K, 1080P, 540P and 270P), 2 QP values (\textsl{i.e.}, 22, 28), and 2 dynamic ranges setting (\textsl{i.e.} high dynamic range and standard dynamic range). Hence, $3 \times 4 \times 2 \times 2 = 48$ Processed Video Sequences (PVS) were generated. We conducted the subjective tests utilizing both the ACR and PC protocol, where 25 participants were involved. In the PC test, the Adaptive Rectangular Design (ARD)~\cite{li2013boosting} was employed to select the comparison pairs for the subject actively, that means his or her playlist was generated based on all previous participants' results. As there are 48 PVS per content, which leads to 48 pairs per reference. To align the scores cross contents, several cross content pairwise comparisons were also included in the test. In concrete words, only the lowest versus highest quality between the HRCs are compared, ending out 6 extract pairs. The viewing distance, environment, \textsl{etc.} were set up according to the ITU standards~\cite{recommendation2004144}. An LG OLED TV was used during the test.

\end{itemize}


%

\subsection{ Experimental results}
\subsubsection{Visualization of EIG} To have a better understanding of EIG, a mesh-grid of EIG versus different $s_{ij}$ and $\sigma{ij}$ is plotted in Figure~\ref{fig:sim} (a). It could be observed that pairs have smaller  $s_{ij}$ and higher $\sigma{ij}$ (\textsl{i.e.}, higher uncertainty) are of higher information. This observation is aligned with the study summarized in~\cite{silverstein1998quantifying}.

\subsubsection{Results on simulated data}
\begin{figure}[!htbp]
\centering
 \mbox{ \parbox{1\textwidth}{
  \begin{minipage}[b]{0.20\textwidth}
   \subfloat[]
  {\label{fig:figB}\includegraphics[width=\textwidth]{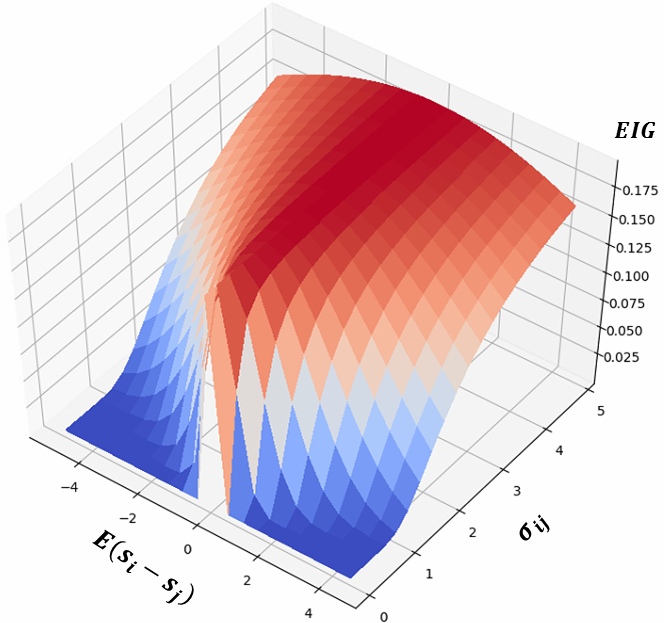}} 
  \end{minipage}
    \begin{minipage}[b]{0.25\textwidth}
   \subfloat[  ]
  {\label{fig:figB}\includegraphics[width=\textwidth]{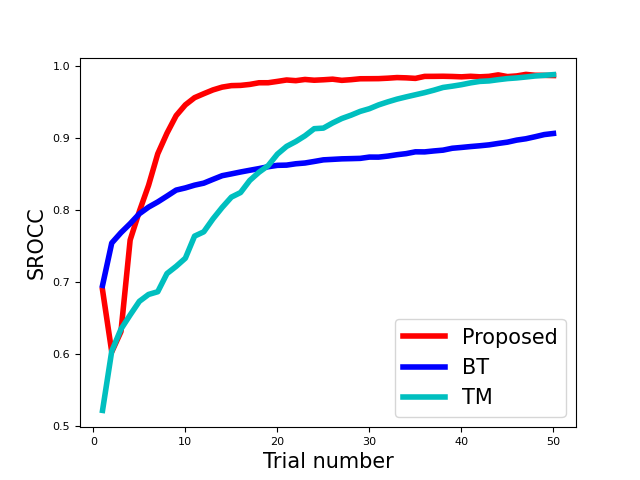}} 
  \end{minipage} 
}}
\caption{(a) mesh-grid plot of EIG regarding $E(s{ij})$ and  $\sigma_{ij}$; (b) performances of considered data on simulated data.}
\squeezeup   \squeezeup
\label{fig:sim}
\end{figure}

The results of the simulation experiment are depicted in Figure~\ref{fig:sim} (b). The performance of the proposed models start to outperform the TM and BT models after the $5^{th}$ trial. The performance of the proposed framework saturate at around 0.97 in terms of SROCC after 15 trials while TM reaches the same performances after 40 trials and the maximum SROCC values of BT is only 0.90. These observations indicate that the proposed framework is of advantage and achieves higher performance with less budgets (\textsl{i.e.,} trials). A better trade-off between the discriminability (performance) and efficiency (budgets) could be achieved.

\subsubsection{Agreement test}

As emphasized earlier in the paper, the conversion from pair comparison preferences to quality rating/ranking scores is important. In order to compare the proposed conversion model with other commonly used models, the agreement test suggested in~\cite{li2011study} was conducted to evaluate the agreement between the converted data and the original ground truth. More specifically, the ground truth $PCM$ is given as input to the under-test pair comparison data conversion model to obtain the recovered rating scores. A matrix $PCM_c$ is then formed by comparing all possible pairs using the converted scores. For each element within the matrix, if the converted score of stimulus $i$ is larger than the one of $j$ then 1 is assigned to $PCM_c(i,j)$, otherwise 0 is assigned instead (\textsl{i.e.,} it is a binary matrix without considering the diagonal elements). Before its comparison with the ground truth matrix $PCM$ that aggregates both the ACR and PC data, $PCM$ is normalized into the interval of $[0,1]$ and transferred to a binary matrix with a threshold 0.5 (\textsl{i.e.,} if lager than 0.5 then set as 1, otherwise set as 0). With the transferred ground truth matrix $PCM_{t}$, it is then compared element-wise to $PCM_c$ to calculate the proportion of elements that has the same values (\textsl{i.e.,} the number of same elements divided by the number of total elements). This \textsl{agreement proportion} quantifies to which degree the recovered scores agree with the ground truth.    

The agreement test results of the proposed conversions model (with Thurstone Case III), BT and TM models on the four real-world datasets are reported in Table~\ref{tab:agr_test}. Overall, it is shown that the rating scores recovered by the proposed conversion model are the most consistent with the observers’ subjective perception on four real-world dataset. 

\begin{table}[!hbpt] 
\begin{center}
\caption{ Agreement proportion of the considered models.   }
\begin{tabular}{|c|c|c|c| } \hline 
Dataset / Model  & TM  & BT & Proposed \\ \hline 
  Kaist &  0.9583 & 0.9614 & \textbf{ 0.9629 } \\ \hline   
  IVC &   0.9584 &  0.9589  & \textbf{0.9602} \\ \hline 
  DIBR &0.9823 & \textbf{0.9829} & \textbf{0.9829}\\ \hline 
  Streaming  & 0.9839 & 0.9848  & \textbf{0.9883} \\  \hline 
\end{tabular}
\label{tab:agr_test}
\end{center}
  \squeezeup   \squeezeup \squeezeup
\end{table}




\begin{figure}[!htbp]
\centering
 \mbox{ \parbox{1\textwidth}{
  \begin{minipage}[b]{0.25\textwidth}
   \subfloat[Kaist dataset ]
  {\includegraphics[width=\textwidth]{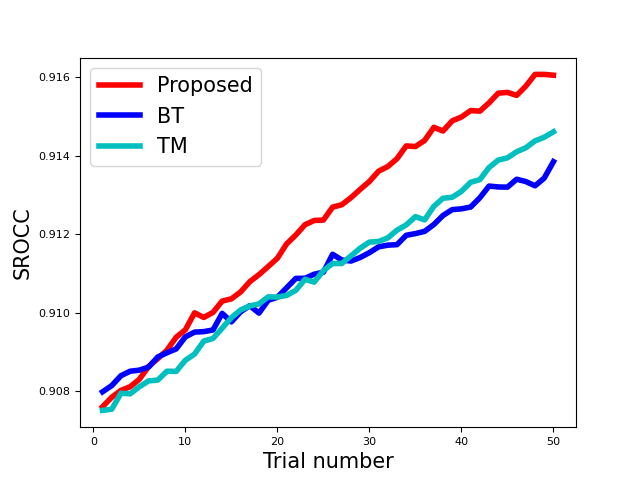}}
  \end{minipage}
    \begin{minipage}[b]{0.25\textwidth}
   \subfloat[IVC image dataset ]
  {\includegraphics[width=\textwidth]{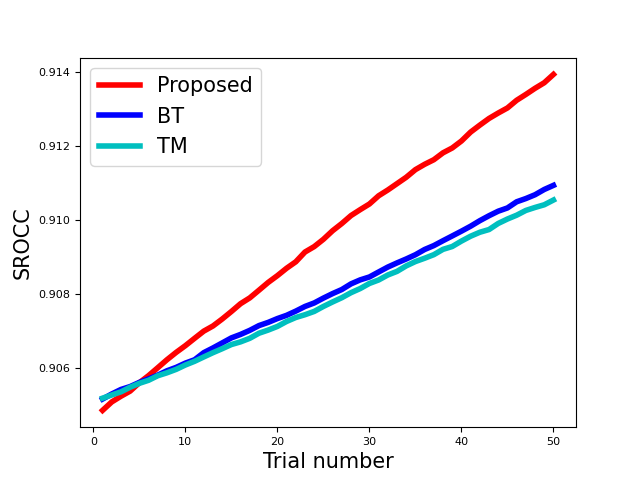}} 
  \end{minipage}  
  \\
    \begin{minipage}[b]{0.25\textwidth}
   \subfloat[DIBR image dataset ]
  {\includegraphics[width=\textwidth]{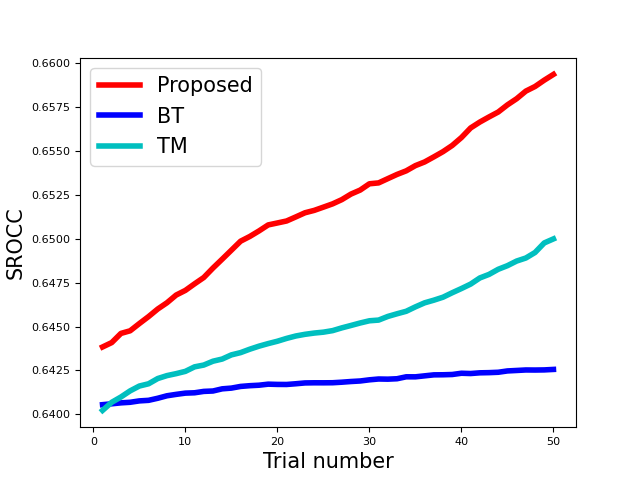}} 
  \end{minipage} 
      \begin{minipage}[b]{0.25\textwidth}
   \subfloat[Streaming video]
  {\includegraphics[width=\textwidth]{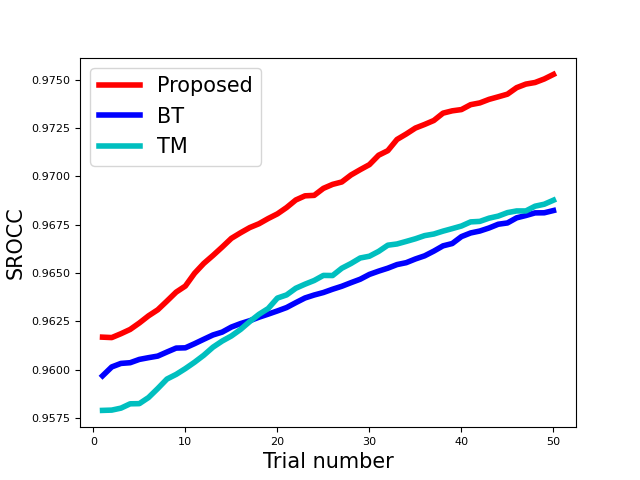}} 
  \end{minipage} 
}}
\caption{Results on real-world datasets.  }
\squeezeup \squeezeup
\label{fig:RW}
\end{figure}
 
\begin{figure}[!htbp]
\centering
 \mbox{ \parbox{1\textwidth}{
  \begin{minipage}[b]{0.25\textwidth}
   \subfloat[Kaist dataset ]
  {\includegraphics[width=\textwidth]{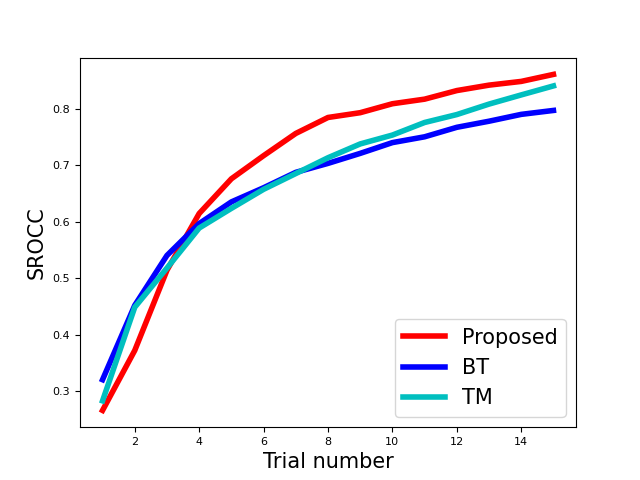}} 
  \end{minipage}
    \begin{minipage}[b]{0.25\textwidth}
   \subfloat[IVC image dataset ]
  {\includegraphics[width=\textwidth]{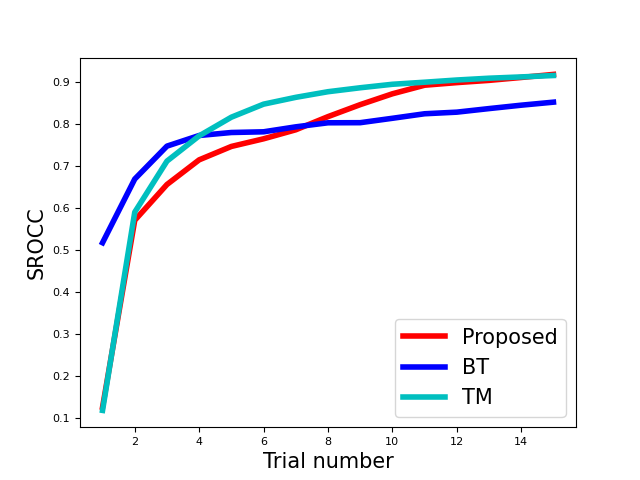}} 
  \end{minipage}  
  \\
    \begin{minipage}[b]{0.25\textwidth}
   \subfloat[DIBR image dataset ]
  {\includegraphics[width=\textwidth]{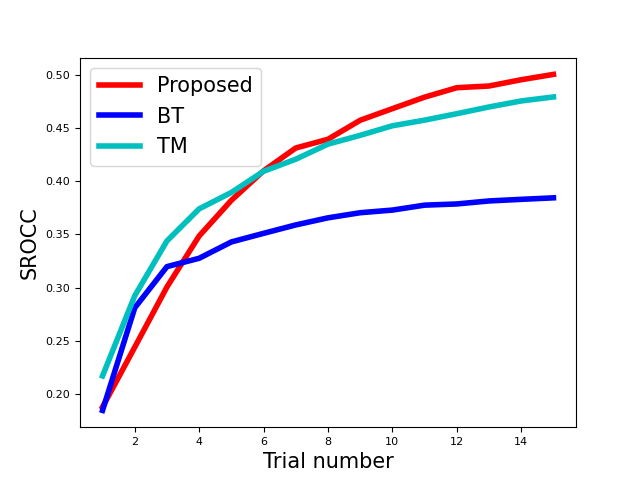}} 
  \end{minipage} 
      \begin{minipage}[b]{0.25\textwidth}
   \subfloat[Streaming video]
  {\includegraphics[width=\textwidth]{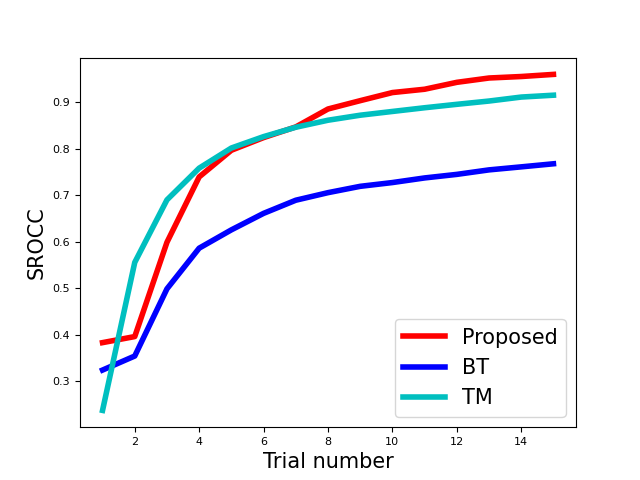}} 
  \end{minipage} 
}}
\caption{Results on real-world datasets without ACR initialization.} \squeezeup \squeezeup
\label{fig:RW_no_ACR}
\end{figure}

\subsubsection{Results on real-world data} 
 
 Figure~\ref{fig:RW} presents the results on the real-world datasets. In general, the maximum SROCC values of the proposed model on the four datasets are superior compared to both BT and TM models. Particularly, our framework starts to outpace the other models after around 10 trials on the Kaist dataset and around 5 trials on the IVC image dataset. Moreover, it outperforms the other models along with trials increase on both the DIBR and the streaming video dataset. It is demonstrated that the performance of the 
active sampling framework could be improved by recovering the variance of stimuli.  
 
To further verify the impact of ACR initialization, we have also conducted experiments without using the ACR initialized PCM matrix. Results are shown in Figure~\ref{fig:RW_no_ACR}. Here, only the results of the previous 15 standard trials (same as done in~\cite{li2018hybrid}) are shown to emphasize the difference of performances in earlier trials without ACR initialization. Compared to Figure~\ref{fig:RW}, it is obvious that the starting performances of all the considered models on the four datasets are significantly worse without considering using the ACR data. For example, the starting SROCC values (\textsl{i.e.,} $1^{th} - 2^{nd}$ trials) of the models in Figure~\ref{fig:RW_no_ACR} (a) are between $ [0.2 , 0.4 ]$, while the ones in Figure~\ref{fig:RW} (a) are around 0.908. It is demonstrated that significant amount of budget could be saved if ACR data is fully used for initialization.

\section{Conclusion}
 In this study, we present a novel active sampling framework to reach a better trade-off between discriminability and efficiency for subjective quality data collection. Within the framework, ACR data is fully exploited for initialization and combined with active sampled pairs comparisons so that budgets could be saved for distinguishing uncertain or similar pairs. In addition, by taking the variance of the stimuli into account, the underlying ground truth quality could be aggregated more accurately. Throughout experiments, the advantages and effectiveness of the proposed framework has been demonstrated.   
 
\bibliographystyle{aaai21}
\balance
\bibliography{ref.bib}

\end{document}